\documentclass[letterpaper, 10 pt, conference]{ieeeconf}

\IEEEoverridecommandlockouts % Only needed to use the \thanks command.
\usepackage{amsmath}                      % Math macros and such
\usepackage{amssymb}                      % Math symbols
\usepackage{algorithm}                    % Algorithm float environment
\usepackage{algorithmic}                  % Algorithm environment
\usepackage[T1]{fontenc}                  % Output non-ascii characters correctly
\usepackage{graphicx}
\usepackage{subcaption}                  % Includes images
\usepackage[urlcolor=blue,colorlinks=true]{hyperref} % For URL links
\usepackage[utf8]{inputenc}               % Enable recognizing utf-8 chars
\usepackage{soul}                         % Latex's best underline command (ul)
\usepackage[usenames,dvipsnames]{xcolor}  % Colors for text coloring
\usepackage{multirow}                     % derp derp

\newtheorem{definition}{Definition}  
\newtheorem{theorem}{Theorem}
\newtheorem{lemma}{Lemma}     % Setup numbered definitions.

\newcommand{\dof}{{\sc dof}}

\newcommand{\cspace}{\ensuremath{\mathcal{C}_{space}}}
\newcommand{\cfree}{\ensuremath{\mathcal{C}_{free}}}

\newcommand{\comp}[1]{$\mathcal{O}(#1)$}

\begin{document}

\title{\LARGE \bf
Representation-Optimal Multi-Robot Motion Planning using Conflict-Based
Search}

\author{Irving Solis$^{1}$, Read Sandstr\"{o}m$^{1}$, James Motes$^{2}$ and
  Nancy M. Amato$^{2}$% <-this % stops a space
\thanks{$^{1}$Irving Solis and Read Sandstr\"{o}m are with the Texas A\&M
University Department of Computer Science and Engineering, College Station, TX,
77840, USA. \{irvingsolis89, readamus\}@tamu.edu.}%
\thanks{James Motes and Nancy M. Amato are with the University of Illinois Urbana-Champaign
Department of Computer Science, Urbana, IL, 61801, USA. \{jmotes2, namato\}@illinois.edu.}%
\iffalse
\thanks{This research supported in part by NSF awards
%EIA-0103742,
%ACR-0081510, ACR-0113971, CCR-0113974, ACI-0326350,
CNS-0551685, %NSF CRI Infrastructure - hydra, rain
%CCF 0702765, %NSF CCF Compilers - Lawrence
%CCF-0833199, %NSF HECURA STAPL
CCF-1439145, %NSF XPS Aero
CCF-1423111, %NSF Bio2
%CCF-0830753, %NSF Bio
IIS-0916053, %NSF ACD
IIS-0917266, %NSF Group Behaviors
EFRI-1240483, %NSF origami
RI-1217991, %NSF FIRM w/ Suman
%by NSF/DNDO award 2008-DN-077-ARI018-02, %SHIELD
by NIH NCI R25 CA090301-11, %NIH Training grant, Ray Caroll
and
by DOE awards
%DE-FC52-08NA28616, %CRASH
%DE-AC02-06CH11357, %CESAR
DE-NA0002376, %CERT
B575363.%DOE LLNL
}% <-this % stops a space
\fi
}

\maketitle
%\Read{I think the title should be a bit more bold and reflect the optimality
%aspect of the work to draw attention. How about 'Roadmap-Optimal Multi-Robot
%Motion Planning using Conflict-Based Search'?}

\thispagestyle{empty} % From IEEE template file.
\pagestyle{plain}     % From IEEE template file.

% Abstract --------------------------------------------------------------------%

\begin{abstract}
	 Multi-Agent Motion Planning (MAMP) is the problem of computing feasible paths for a set of agents given individual start and goal states. 
	Given the hardness of MAMP, most of the research related to multi-agent systems has focused on multi-agent pathfinding (MAPF), which 
	simplifies the problem by assuming a shared discrete representation of the space for all agents. 
	The Conflict-Based Search algorithm (CBS) has proven a tractable means of generating optimal solutions 
	in discrete spaces. However, neither CBS nor other discrete MAPF techniques can be directly applied to solve MAMP problems because the 
	assumption of the shared discrete representation of the agents' state space.
	In this work, we solve MAMP problems by adapting the techniques discovered in the MAPF scenario 
	by CBS to the more general problem with heterogeneous agents in a continuous space. We
	demonstrate the scalability teams of up to 32 agents, demonstrate the ability to handle up to 8 high DOF
	manipulators, and plan for heterogeneous teams. In all scenarios, our approach plans significantly faster
	while providing higher quality solutions.

\end{abstract}

% Content ---------------------------------------------------------------------%
\section{Introduction}

In automated manufacturing, high DOF manipulators working in tight coordination
must avoid colliding with each other while planning efficient motions.
Heterogenous teams used
on construction sites or in search-and-rescue missions must similarly coordinate collision free motions.
Video games with large multi-agent teams must also produce feasible plans.
These are only a few examples
of the multi-agent motion planning (MAMP) problem.

MAMP solutions require geometrically feasible collision free paths through continuous state spaces for each agent in the team.

There are two standard approaches to MAMP: coupled and decoupled. Coupled methods are able
to provide optimal solutions with respect to the state space representation but search over the joint state space. Decoupled
methods search over the individual state spaces of each robot but cannot guarantee completeness or optimality as
they explore individual agent state spaces in isolation before combining them later.
Hybrid approaches try to leverage the benefits of both approaches.

\begin{figure}[h!]
	\centering
	\begin{subfigure}[b]{0.45\linewidth}
		\includegraphics[width=\linewidth]{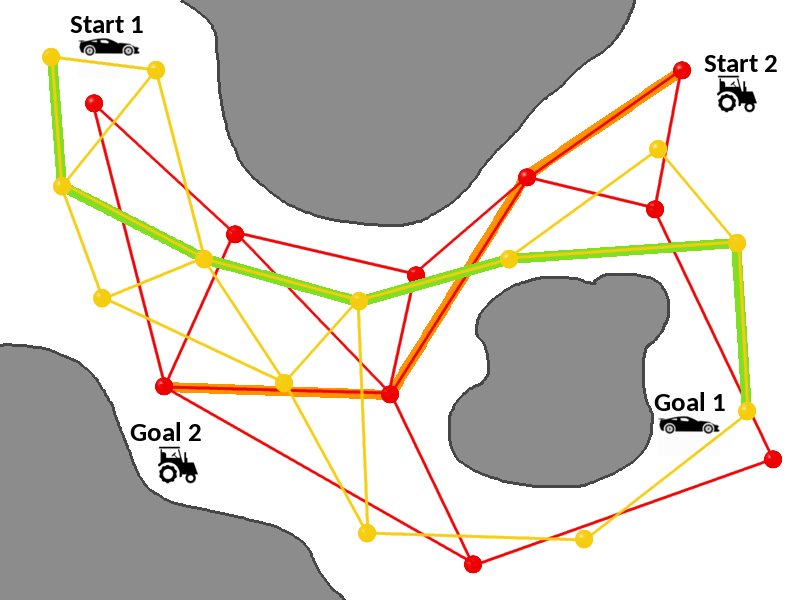}
		\caption{}
	\end{subfigure}
	\begin{subfigure}[b]{0.45\linewidth}
		\includegraphics[width=\linewidth]{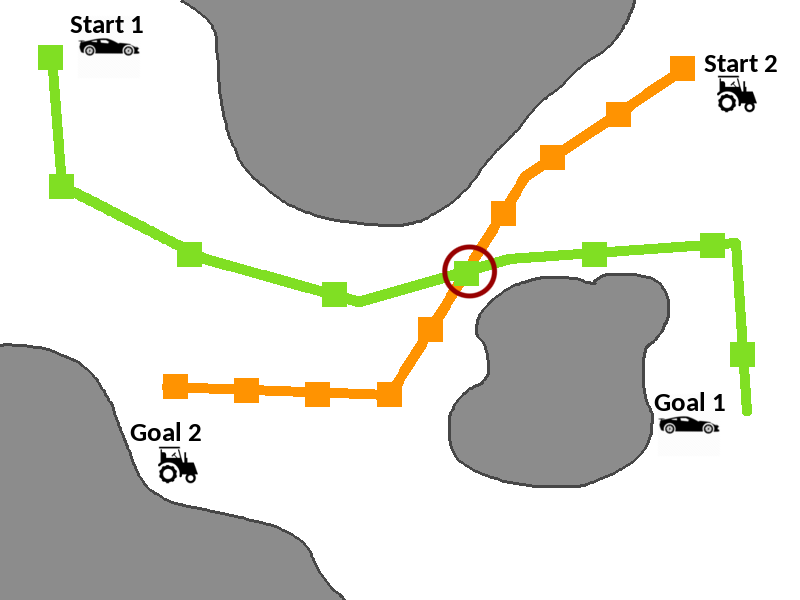}
		\caption{}
	\end{subfigure}
	\begin{subfigure}[b]{0.45\linewidth}
		\includegraphics[width=\linewidth]{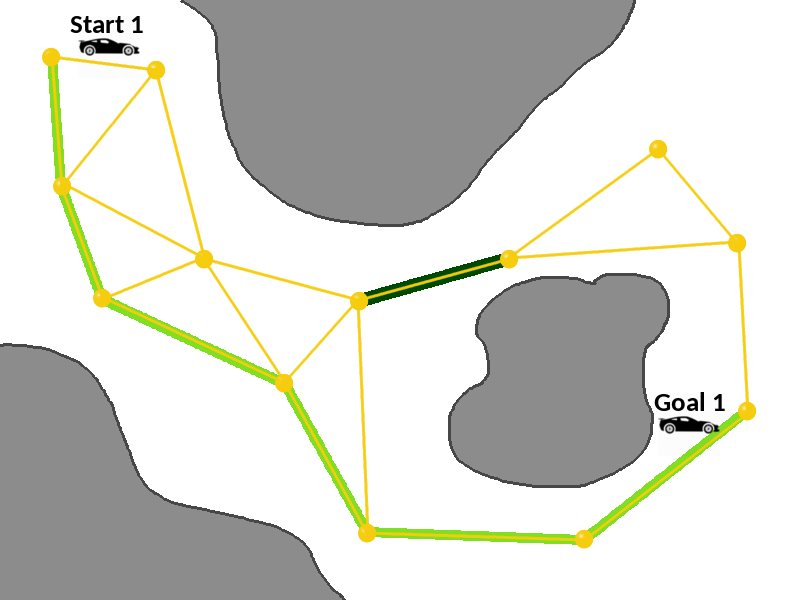}
		\caption{}
	\end{subfigure}
	\begin{subfigure}[b]{0.45\linewidth}
		\includegraphics[width=\linewidth]{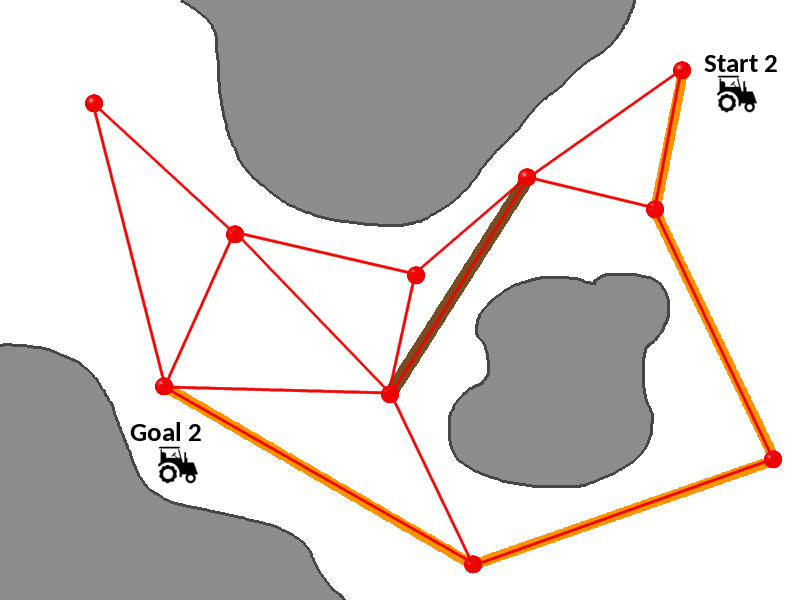}
		\caption{}
	\end{subfigure}
	\caption{(a) Individual roadmaps and paths are computed separately. (b) Paths are discretized into uniform motion time segments, then inter-agent collision detection is achievable. (c,d) The conflict found is mapped back to the corresponding edges, then new collision-free can be computed. }
	\label{fig:roadmaps}
\end{figure}

Given the complexity of MAMP, most of the research related to multi-agent systems has focused
to solving an easier subproblem, multi-agent pathfinding (MAPF). The pathfinding subproblem operates
in a discrete state space as opposed to the continuous space considered in the motion planning problem.
However, most real world problems, such as the high DOF manipulator and heterogenous multi-agent
teams mentioned previously cannot be solved directly by discrete MAPF techniques
as their conflict detection mechanism depends
on a unified representation of the agents' state space.

In this work, we present an efficient method for solving MAMP problems with
heterogeneous agents in a continuous state space.
We generalize a recent efficient and optimal MAPF technique, Conflict Based Search 
(CBS) [27], to continuous state spaces.
Our new MAMP method, which we call CBS-MP, 
produces conflict-free motion plans
for a heterogeneous multi-agent team which are optimal with respect to the 
individual agent state space representation.

We show our new method, CBS-MP, leads to improved performance in comparison to both standard coupled and decoupled PRM variants. We showed an increased 
scalability, planning for up to 32 agents while the PRM variants were unable to plan for 16 agents. We planned for up to 8 high DOF 
manipulators over 60 times faster on average than the fastest PRM variant run. Additionally, we demonstrate the flexibility of the 
approach on a heterogeneous problem with a manipulator, mobile robot, and aerial robot operating in the same workspace.
In addition to significantly improved planning times, we show lower cost solutions than the coupled and decoupled PRM variants.

%------------------------------------------------------------------------------%

\section{Related Work}
We first examine recent work on multi-agent pathfinding and then
look at the problem in the motion planning domain.

\subsection{Multi-Agent Pathfinding}

In recent years, multi-agent pathfinding (MAPF) has received considerable attention.
The problem is defined by a set of agents, a graph, and a corresponding start and goal location
for each agent. A solution consists of a set of collision free paths across the graph transitioning each agent from
the start to the goal. The quality of solutions are traditionally measured by either
the \textit{sum-of-cost}, summed cost of all paths, or \textit{makespan}, maximum individual path. An optimal
solution minimizes the desired metric.

Generally, all MAPF methods can be classified into three major groups: coupled,
decoupled, and hybrid.
\subsubsection{Coupled Approaches}
In coupled approaches, all of the agent paths are
computed in unison, step by step. These techniques work in a joint space of all agent
states. They tend to provide stronger guarantees on feasible paths and
minimum cost because they explore the joint space, however,
this has a high computational cost as the
dimensionality of the joint \cspace increases with the number of robots and
their degrees of freedom.

A simple MAPF solver can be implemented by concatenating all the agents into
one single state. A single search algorithm like A* can then be used to
traverse the joint space of all agents to get to the state solution ~\cite{hnrb-afbfthdomc-68}.
Unfortunately, since the joint-space is the result of the cross-product of all
the single spaces, the number of states grows exponentially with respect to the
number of agents (i.e. A* in joint space is \comp{V^{2n} E \lg E} for $n$
agents). As a result, this type of approaches is only practical for a small
number of agents.

In order to deal with this issue and speed up the search,
other works attempted to prune the search by using some heuristics and expand
fewer nodes than regular A* ~\cite{fgssbssh-peawsnd-12,gfssshs-epea-14,k-dfidaoats-85}. However, deterministic
algorithms such as A* may take a significant amount of time to solve a
problem, or sometimes are not even able to solve it. Alternative approaches have
modeled the multi-agent pathfinding problem as Integer Linear Programming (ILP)
and Boolean Satisfiability (SAT) problems. In ~\cite{yl-omppogcaaeh-16} Yu et. al.
mapped the problem to a network-flow and applied ILP algorithms to optimally
solve four objectives: the makespan, the maximum distance, the total arrival
time, and the total distance. In ~\cite{ekos-agfffppwma-13} the multi-agent pathfinding problem
is recast as a SAT problem where logical variables are used to represent time
and physical locations for each agent as well as for obstacles. Then logical
rules are enforced to solve the problem while avoiding collisions and satisfying
other constraints.

However, all those techniques suffer from being susceptible to increase their
computational cost as the number of robot increases or even when dealing with a
few robots with high \dof\hspace{0.4mm} since they navigate in the joint
state space.

\subsubsection{Decoupled Approaches}
To mitigate this potential computational cost,
attention has turned to decoupled algorithms. Instead of planning all paths
in unison, each path is planned individually. Once all paths are
computed, they are adjusted according to defined priorities in order to
avoid inter-agent collisions. Decoupled approaches work in lower-dimensional
spaces allowing these strategies to rapidly compute feasible paths
for problems with a large number of agents. However, these techniques
explore the state space of individual agents in isolation and later combine the solutions.
This leads to difficulties with completeness and optimality, and there is often
the possibility of failing to find feasible paths on solvable problems.

To achieve the group coordination of multiple
robots, ~\cite{cgg-pocomrwsg-12} computes a set of pre-specified paths over a roadmap, considering
motion safety and minimizing the traveling time. Then, possible collision points
of the local paths of the robots are estimated. Finally, for each robot path,
a fixed velocity is computed in order to avoid all conflict points. In
~\cite{cnsmfv-adppfcimrs-13,dh-dppfmatwcc-12,rl-cmrppbhpa-06,bo-pmpmr-05,apv-bcp-06}, priorities are
assigned to each agent based on different heuristics, and then their individual paths are computed in
decreasing priority order while treating higher priority agents as dynamic
obstacles.
%~\cite{ms-mrppapcuga-17} presents a genetic algorithm implementation where the optimal individual paths serve for both obtaining the priority order of the robots and getting a fitness function for a path genetic mutation. Then, when a conflict is detected, the highest priority path is mutated several times until the conflict gets resolved. Then the resulting path is added as a dynamic obstacle.
Unfortunately for decoupled approaches, given the drawback of limiting the joint-space exploration, decoupled planners are
\textit{incomplete}. %They may fail to find a solution when one exists.

Not all decoupled approaches suffer from that issue. CBS~\cite{ssfs-cbsfomap-15} alleviates
the incompleteness issue by considering all the possible ways
in which conflicts between individual paths can be resolved. This is done by using a two-level
search. The low-level generates individual robot paths in a decoupled manner. The high-level
detects conflicts in these paths and generates constraints for the the low-level search to modify
the individual paths. This is the method
we extend to multi-agent motion planning in this paper.

Recent improvements and variations on the CBS algorithm have been proposed. In
~\cite{bfss-dstwiobcimap-15}, \textit{ByPass-CBS} allows internal modifications to conflicts
found in the high-level search that compute alternative paths without expanding the high-level search.
\textit{Continuous-Time-CBS} ~\cite{ayas-mapwct-19}
extends the state-space search to plan in
continuous-times.

\subsubsection{Hybrid Approaches}

Due to the tradeoff between faster computation times and finding
optimal cost solutions, researchers explored new ways of leveraging the
strengths of both coupled and decoupled techniques. These techniques are also
known as hybrid approaches. In ~\cite{wc-sefmpp-15}, \textit{M*} solves the MAPF
problem by initially planning in a fully decoupled manner. Then when a
inter-robot conflict arises, the individual search is backtracked until the last
collision-free state, at which point, the conflicting agents are joined into a coupled
meta-agent, and new collision-free paths are computed by using a coupled
planner for those agents. In the worst of the cases, if all agents are in
collision at the same place and time, M* becomes a fully coupled planner.

%Recent extension to the CBS algorithm also use this meta-agent concept.
%In ~\cite{ssfs-macbsfomapp-12}, \textit{MetaAgent-CBS}, when a conflict is found, rather
%expanding the high-level search, the conflicting agents are merged into a
%\textit{meta-agent} and coupled planner is used in the low-level search to compute its path.
The meta-agent concept in leveraged in \textit{MetaAgent-CBS}, a recent extension to CBS.
when a conflict is found, rather
expanding the high-level search, the conflicting agents are merged into a
\textit{meta-agent} and coupled planner is used in the low-level search to compute its path.
In \cite{bfsstbs-iicbsafmap-15}, an improved version of \textit{MetaAgent-CBS} is proposed, where the
decision of merging agents has only local effects and conflicting agents are
treated again as single agents in further high-level exploration.

\subsection{Multi-Agent Motion Planning}

Motion planning is the problem of finding a feasible path between a start and goal
in a continuous state space that is usually intractable to represent explicitly.
This is a superset of the pathfinding problem that searches over a discretized state space.
In motion planning, the state space is comprised of the set of all
possible agent configurations known as the \textit{configuration space} (\cspace)~\cite{lw-apcfpapo-79}. A solution to
the motion planning problem is a continuous path in a subset of \cspace called \textit{free space} (\cfree)
consisting of valid configurations.

In response to the complexity of motion planning~\cite{r-cpmpg-79}, sampling-based
motion planners were developed as an efficient means of discovering valid paths
in \cfree. These methods, such as the Probabilistic Roadmap Method (PRM)~\cite{kslo-prpp-96}
attempt to create a \textit{roadmap}, or graph, approximating \cfree. Paths are found by querying
this roadmap.

Not much work has been proposed for
sampling-based multi-agent motion planning (MAMP).
Much of the work
that has been done modifies or extends two widely used single agent sampling
based motion planning algorithms, PRM~\cite{kslo-prpp-96} and RRT~\cite{l-rrtntpp-1998}. In
~\cite{cp-prrtmrs-02} ~\cite{fks-rwr-06} optimizations and improvements to regular RRT are proposed to
enable it to solve multi-agent problems in \textit{Joint-}\cspace. In
~\cite{mnvp-marsbcp-13}, the multi-agent version of RRT* (MA-RRT*) searches for the
shortest path in a group graph that represents the joint-state space of all
agents. The returned solution is then a collision-free joint plan containing
a path for each agent. In ~\cite{Sanchez-2002} ~\cite{sl-uppccdpmrs-2002}, PRM is used to compare the
performance of centralized and decoupled planning for multi-robot systems.
In ~\cite{si-mrmpbic-06}, a coupled planner called \textit{MRP-IC} solves the
MAMP problem in an incremental way. First, an individual path for the first
agent is found and this serves as the basis for recursively guiding a
coupled search in an incremented state-space to find the path of the next
\textit{k+1} agent by reusing the paths of the first \textit{k} agents which
were already computed.

In ~\cite{ksb-tudmptacp-12}, several ideas of how discrete MAPF techniques can be
adapted to continuous problems are provided. "Relocating Roadmap
Nodes", "Merging nodes" and "Addressing Node/Node and Node/Edge
interactions" are some of these ideas, however no experimentation is
provided to validate their applicability.

%------------------------------------------------------------------------------%

%------------------------------------------------------------------------------%
\section{Conflict-Based Search in Continuous Spaces}

We will first review the MAPF version of CBS. We will then explain how
it can be adapted to solve continuous space MAMP problems with
sampling-based motion planning.

\subsection{Review of Conflict-Based Search in Discrete Space}

Conflict-Based Search (CBS) is an optimal decoupled MAPF method~\cite{ssfs-cbsfomap-15}.
The approach utilizes a low-level search to plan individual agent paths and a high-level search
to resolve conflicts between the agent paths.

The low-level search in performs an A* search over a state space
consisting of location and time. The locations
correspond to vertices on a grid roadmap shared by all agents. All vertices are one unit
distance from their neighbors, and moving between neighboring vertices constitutes one timestep.

The high-level search utilizes a Conflict Tree (\textit{CT}). Nodes in this tree contain a set of paths
for all agents, a set of path constraints for the various agents, and a cost of the solution. The
set of initial paths planned for all agents individually is used to create the root node. The root
node contains no constraints, and the cost is relative either the \textit{SOC} or \textit{makespan} for the solution.

\subsubsection{Conflict Detection}
Solutions in \textit{CT} nodes are evaluated for path conflicts. A conflict occurs when
two agents share the same graph vertex at the same timestep and is defined by the
graph vertex, the timestep, and the agents involved $c=<v,t,a_i,a_j>$. The conflict detection
process stops at the earliest conflict in the solution.

\subsubsection{Conflict Resolution}
Whenever a conflict $c$ is detected, a path constraint is generated for
each agent involved. A constraint for an agent$i$ consists of the time $t$ of the collision,
and the vertex $v$. A child \textit{CT} node is generated
for each constraint. Each child node additionally copies all of the constraints of the parent
node.

The low-level
planner then computes new individual paths which are consistent with the current
conflicts sets. In the MAPF problem, all agents lie on the same
graph, and a constraint can be easily mapped to an "invalid" state to be avoided
during the low-level search.

The search continuous evaluates the next lowest cost unexplored node in the \textit{CT}.
The search stops upon finding the first conflict free or goal node. As all remaining unexplored nodes
contain solutions at least as expensive as the goal node, the solution in the goal node is returned
as the optimal solution.

\subsection{Adapting CBS to Sampling-based Motion Planning}

In general planning problems, heterogeneous agents operate in distinct continuous spaces.
Our approach, CBS-MP, adapts the MAPF CBS method, which assumes a shared discrete representation, 
to this more general scenario. 

Our adaptation starts by 
sampling individual roadmaps for each agent using standard decoupled PRM techniques, and then 
uses the two-level method of CBS to query the roadmaps.
MAPF CBS is unable to handle this discretization because the representation is disjoint and the
length and duration of the edges are nonuniform. These issues manifest themselves during 
the conflict detection and resolution stages of CBS. 

When the individual agent paths are checked 
for conflicts the paths are further discretized into uniform time resolution segments. 
These segments can be of different lengths in the roadmaps of different agents, but 
each segment takes the same time duration for the corresponding agent to traverse.
This allows the sequence of segment endpoints along each agent's path to be synchronized.
These endpoints correspond to configurations for the associated robots. The configurations
are checked for conflict with standard motion panning collision detection methods.

When a conflict is detected, it consists of a pair of endpoint configurations $p_i,p_j$ and a timestep $t$, $c=<t,p_i,p_j>$.
The constraints generated from a conflict of this form consist of an agent $i$, timestep $t$, and the other agent's
conflicting configuration $p_j$. Motion plans for individual agents satisfying constraints of this form must avoid collision 
with the constraint configuration at the corresponding timestep.
Doing this involves performing a large
number of collision detection calls, so to alleviate this issue, we
first verify if the edge we are checking during graph search contains the
timestep of the constraint object, and if so, we perform the collision
detection against the other agent configuration. Otherwise we proceed with
the search.

\subsection{Size-Limited Conflict Trees}

In discrete MAPF, it is assumed that the given state space representation is complete, and thus all possible paths
can be found by searching over it. In the continuous space motion planning problem this is not the case. As the
roadmaps only contain a sampling of the state space, there are more feasible paths not contained within the current roadmaps.

In instances where large numbers of conflicts are found, it is often useful to further expand the roadmaps as opposed to spending
the computation time on resolving large \textit{CT} trees. To do this, we allow a maximum \textit{CT} size parameter that limits the
number of nodes explored in the \textit{CT}. Upon reaching this maximum size, CBS-MP returns to the sampling phase of the
method, and grows the individual agent roadmaps before starting the query phase and \textit{CT} exploration over again.

Finite \textit{CT}s are not necessarily
complete, as problems requiring more complex coordination would not be
discovered. However in the motion planning setting, the roadmaps can be improved
if the query fails, thus opening new paths to attempt. This causes the planner
to search for paths which meet the maximum complexity requirement by continually
refining the roadmaps to discover one. Searching \textit{CT}s with no maximum size
resembles the discrete pathfinding CBS with the assumption that the initial roadmaps are able to provide a
desired solution without further sampling.

%------------------------------------------------------------------------------%

\subsection{Theoretical Properties}

In this subsection, theoretical details about CBS-MP are discussed. 
We start by formally defining the Multi-agent Motion Planning problem 
and what constitutes a solution for it.

\begin{definition}
	A \textit{multi-agent motion planning (MAMP) problem} consists of an
	environment $E$, set of agents $A$, and initial and goal conditions for each
	agent $a_i \in A$. Let the free configuration space for agent $a_i$ be
	denoted as $\mathcal{C}_i$. Then for each agent $a_i \in A$ the initial
	state is a configuration $s_i \in \mathcal{C}_i$ and the goal condition is
	reaching a set $G_i \subset \mathcal{C}_{i}$.
\end{definition}

\begin{definition}
	Let $P$ be a set of paths $\rho_i(t) : [0, t_{final}] \rightarrow C_i$ for
	each agent $a_i \in A$. We say that $P$ is a solution to an MAMP problem if
	for all times $t \in [0, t_{final}]$ there are no collisions between the robot
	configurations $\rho_i(t)$.
\end{definition}

\begin{definition}
	Let $P$ be a solution to an MAMP problem extracted from a set of roadmaps
	$r_i \in R$ for each agent $a_i \in A$. $P$ is said to be
	\textit{representation-optimal} with respect to $R$ and a cost metric
	$C(R, P) = C*$ if there is no set of valid paths $P'$ in $R$ with lower cost
	$C(M, P') < C*$.
\end{definition}

\begin{definition}
	A \textit{representation-optimal} team solution is optimal with respect to the
	current representation of each individual agent's state space.
\end{definition}
Many MAPF methods claim to provide optimal solutions. This optimality is with respect to their input
state representation or roadmap which is assumed to be perfect. While some sampling-based
motion planning methods provide asymptotic optimality, this never achieves true optimality in
any real application. In this work, we focus on providing a team path which is optimal with respect to the
current representation of each individual agent's state space.

\begin{lemma}
	\textbf{Termination:} The CBS-MP query phase will terminate in finite time.
\end{lemma}

In the worst case, the CBS-MP query phase will explore all possible paths encoded in $R$ which
do not include cycles or waiting. The number of paths is thus finite, and CBS-MP
will at most explore each of a set of finite paths. Hence, CBS-MP will terminate
in finite time.

\begin{lemma}
	\textbf{Validity:} If the CBS-MP query finds a solution, it will be collision-free.
\end{lemma}

CBS-MP maps the agents' heterogeneous representations to a common representation
with uniform time discretization. For any common representation where conflicts
can be detected, the inverse map can be used to assign the conflict to the
associated state in the agents' individual representations. These are the
generalized requirements for conflict resolution in CBS, and any produced
solution will thus be free of conflicts.

\begin{lemma}
	\textbf{Completeness:} If a solution without cycles or waiting exists in $R$,
	the CBS-MP query will find it.
\end{lemma}

As indicated previously, in the worst case the CBS-MP query will explore all possible
paths in $R$. On discovery of the first valid path $P$, no conflicts will be
detected by collision checking with uniform time discretization, and $P$ will be
identified as a solution.

\begin{lemma}
	\textbf{Optimality:} If a \textit{representation-optimal} path without cycles
	or waiting exists in $R$, the CBS-MP query will find it in a finite time.
\end{lemma}

The individual agent paths are generated with Dijkstra's algorithm on a fixed
state space representation with strictly non-negative weights, which represent
motion time. The individual costs will thus increase monotonically on each
replan and are admissible. For any admissible group cost metric, the CBS-MP query
explores group paths in best-first cost order. A discovered solution will
therefore have the best possible cost of those encoded by $R$. By the
\textit{completeness} lemma, we will find such a path in finite time if it
exists.

\begin{theorem}
	\label{theorem1}
	By lemmas 1,2,3,4, the CBS-MP query reduces to CBS with a different conflict resolution
	mechanism. It thus terminates in finite time, is complete, and provides a
	representation-optimal solution.
\end{theorem}

We now discuss the implications the properties of the CBS-MP query has on the full motion planner.
\begin{lemma}
	Each time we run the query, we will find the representation-optimal path if it exists.
	\label{lemma:query}
\end{lemma}
By Theorem~\ref{theorem1}, the query will return the representation-optimal path for whatever the 
current roadmap representation when the query is run.
\begin{lemma}	
	If a solution exists, we will find roadmaps that encode it eventually.
	\label{lemma:prob-complete}
\end{lemma}
Due to probabilistic completeness of the decoupled PRM used to 
generate the roadmaps in CBS-MP, the set of roadmaps $R$ will eventually 
contain all possible paths. 
\begin{theorem}
	The full planner is probabilistically complete and representation-optimal.
\end{theorem}
As any possible solution will eventually be found by Lemma~\ref{lemma:prob-complete}, and
the query will return the representation-optimal path for the current roadmaps by Lemma~\ref{lemma:query}, 
the full planner, CBS-MP,  is probabilistically complete and representation-optimal.

\begin{figure}[h!]
	\centering
	\begin{subfigure}[b]{0.49\linewidth}
		\includegraphics[width=\linewidth]{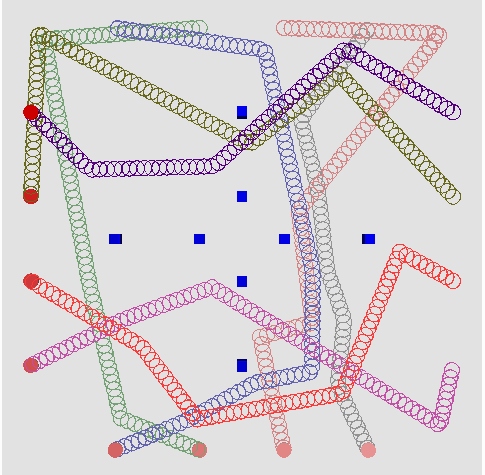}
		\caption{Scaled multi-agent team environment.}
		\label{fig:scenario1}
	\end{subfigure}
	\begin{subfigure}[b]{0.49\linewidth}
		\includegraphics[width=\linewidth]{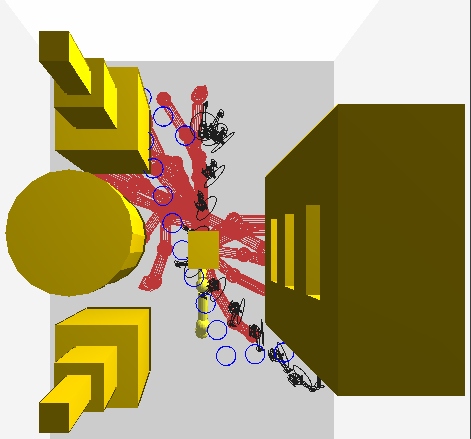}
		\caption{Heterogeneous multi-agent team environment.}
		\label{fig:heter}
	\end{subfigure}
	\caption{a) The agents on the bottom must find paths to the top, and the agents on the left
		must find paths to the right. The paths for each agent making up the solution are
		shown in the various colors.
		(b) A crane occupies the central region of the space while a ground-based and aerial robot must navigate through the same space.}
	\label{fig:manip}
\end{figure}
\section{Validation}

We evaluate our method against the canonical coupled and decoupled PRM.
For our experiments, DecoupledPRM was implemented as
a dynamic-obstacle-based approach and computes individual agent paths incrementally. 
As the coupled approach provides an optimal solution,
and the decoupled approach an improvement in performance, we compare against the characteristics
of both methods.

\begin{figure*}
	\centering
	
	\includegraphics[width=6in]{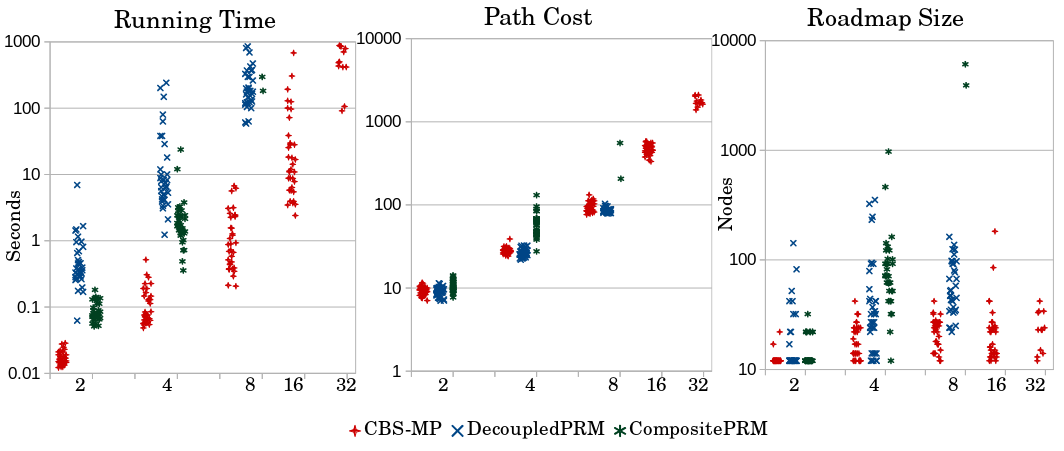}
	\caption[]
	{Scenario I results: These results show the scalability of the method. The X-axis denotes the number of robots. CBS-MP shows faster running times while using smaller roadmaps. CBS-MP is the only method that solved problems with 16 and 32 robots. }
	\label{fig:scal-results}
\end{figure*}

\begin{figure*}
	\centering
	
	\includegraphics[width=6in]{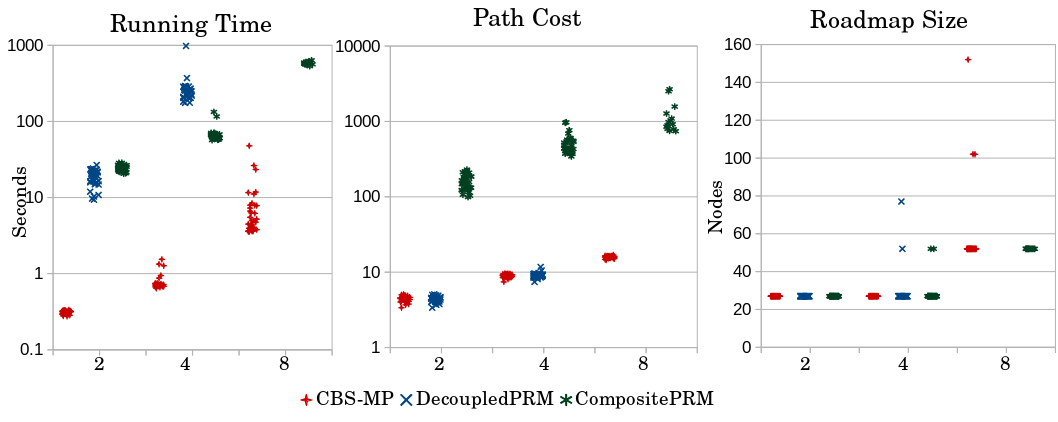}
	\caption[]
	{Scenario II results: These results show the performance of the methods for high-DOF agents. The X-axis denotes the number of robots. CBS-MP shows faster running times. CBS-MP is the only method that solved the problem with eight robots completely. }
	\label{fig:man-results}
\end{figure*}

\begin{figure*}
	\centering
	
	\includegraphics[width=6in]{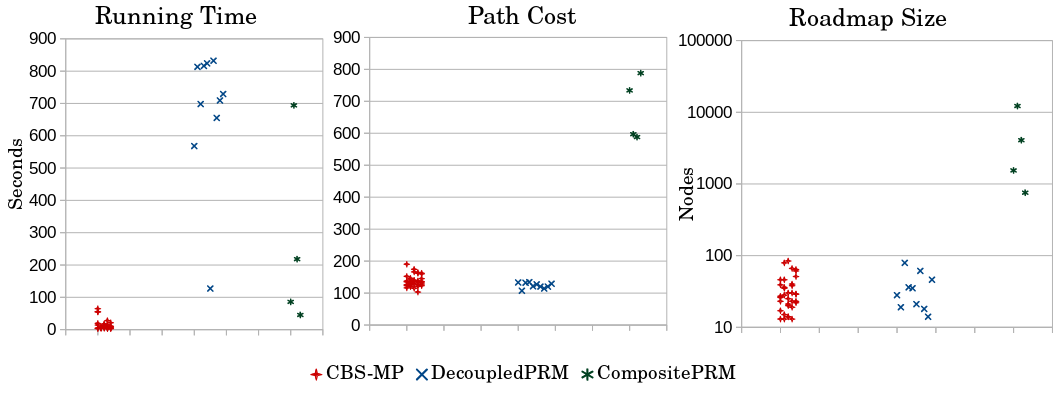}
	\caption[]
	{Scenario III results: These results show the adaptability to heterogeneous systems. CBS-MP shows faster running times. CBS-MP is the only method that solved the heterogeneous-system problem completely.}
	\label{fig:het-results}
\end{figure*}

\begin{figure*}
	\centering
	\includegraphics[width=6in]{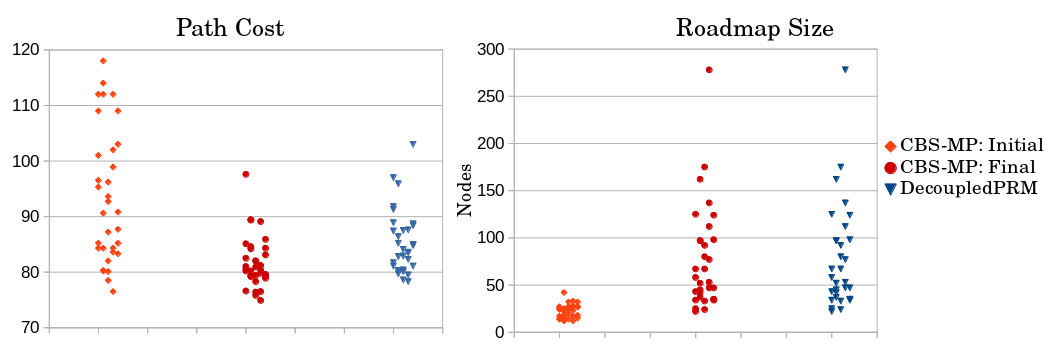}
	\caption[]
	{Scenario IV results: These results show that CBS-MP finds a path with smaller roadmaps than DecoupledPRM. The CBS-MP solution improves as the roadmaps expand to the point needed by DecoupledPRM to find a solution.
	}
	\label{fig:fourth-results}
\end{figure*}

We verify the performance of our method in three main scenarios, large sets of agents, agent teams
with high \dof, and heterogeneous systems. Additionally, a fourth scenario is
included, which studies a head-to-head comparison of the two decoupled methods,
DecoupledPRM and CBS-MP, to explore in motion planning terms, the difference with respect to the node and
edge generations in individual roadmaps.
Agents are not allowed to perform wait or looping actions. Also, they are always
physically present during the whole group plan (i.e. an agent never disappears even if it finishes its plan
before to the rest of the group) and they start their plans simultaneously. CT trees are size-limited to 64 nodes.
All methods are given 1000s seconds to plan at which the attempt is considered a failure.

\subsection{Scenario I}

This scenario evaluates the scalability with respect to the number of agents for each 
methods when a large number of planning conflicts will occur. 
It consists of an open
environment with simple box-shaped agents. Half of the robots move
from the left to the right side and the other half go from the bottom to the top
side (Fig.~\ref{fig:scenario1}). This forces half of the paths to be orthogonal to the other half creating a problem
full of potential inter-robot collisions.

Besides simply analyzing the scalability of the three
methods, this scenario studies the performance of our method when its conflict
tree grows significantly. Using simple boxed-shape robots inside a free environment
without obstacles minimizes extraneous factors that may affect the performance
of the planners. It is important to note that when increasing
the number of robots, we proportionally increased the size of the environment. This allows
us to have the same density per robot in all the tests
ensuring this has no effect on the performance of the planners.

With small numbers of agents, the CompositePRM is able keep relative low planning times,
taking roughly three times as long to plan as CBS-MP and planning faster than DecoupledPRM.
For eight agents, the coupled approach becomes drastically more expensive as the
state space grows too large. Despite the optimal guarantees of the composite approach, 
the solutions are often much more expensive than the other two methods. This is due to 
CompositePRM exploring the joint \cspace of the multi-agent team. 
As the size of the joint \cspace grows exponentially with the number of robots, the probability of sampling joint configurations that are useful relative the individual agent paths becomes increasingly low.

DecoupledPRM scales better, but still takes significantly longer to plan 
as the number of agents increases  compared to CBS-MP(Fig.~\ref{fig:scal-results}). 
Neither method is able to solve in the allotted time for 16 or 32 agents.

The DecoupledPRM method often produces higher quality plans than CBS-MP. This is expected as CBS-MP is
able to find solutions with sparser roadmaps as indicated by the average number of nodes in the agent roadmaps.
DecoupledPRM continues to sample and build denser roadmaps until it finds a solution as opposed to resolving conflicts
within the existing roadmap as CBS-MP does. 

As CBS-MP is representation-optimal, it will always produce a plan at least as good as 
DecoupledPRM on the same roadmap. This is expanded further in Section~\ref{sec::scen4}.

Additionally, while we saw a steady improvement in running time for CBS-MP, we
do not expect this to hold in all cases. Although
unlikely, it is possible for the DecoupledPRM to solve faster. For example, consider an initial roadmap generated that contains
a solution but also many conflicts. CBS-MP will grow the conflict tree attempting to resolve the conflicts, meanwhile, DecoupledPRM
will sample again and may find a path with only a second iteration of roadmap growth. This can result in a faster running time than
CBS-MP. We did not observe this in our experiments.

\subsection{Scenario II}

\begin{figure}[h!]
	\centering
	\begin{subfigure}[b]{0.49\linewidth}
		\includegraphics[width=\linewidth]{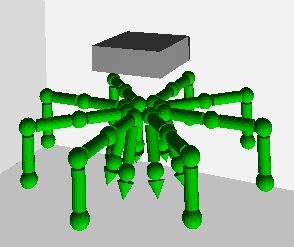}
		\caption{The start configuration.}
	\end{subfigure}
	\begin{subfigure}[b]{0.49\linewidth}
		\includegraphics[width=\linewidth]{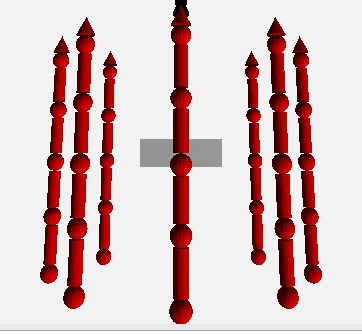}
		\caption{The goal configuration.}
	\end{subfigure}
	\caption{The set of manipulators starts in a convoluted configuration requiring tightly coupled maneuvering
		to reach the goal configuration. }
	\label{fig:manip}
\end{figure}

This experiment shows how our algorithm performs problems with multiple
high-\dof\ robots moving through a shared space. In this second scenario,
several robotic arms start in a confined initial position where each arm is
nearly in contact with all the rest of the robots. This requires a
coordinated approach to transition from the group start configuration to the goal
configuration (Fig.~\ref{fig:manip}). The conflict tree was limited to sixteen nodes to favor simpler
solutions because there are quite a few ways for the manipulators to become
entangled. CBS-MP performs well in this problem even with eight robots(Fig.~\ref{fig:man-results}).Due to its greedy-priority nature, DecoupledPRM always picks the shortest path for one of the manipulators, which will usually go through the highly contested middle space. This decreases the probability that the others will find a path because the majority of paths require their distal joints to use parts of the middle space. As the number of robots increases, so does the density and the level of congestion in the middle. This prevents it from finding solutions where CBS-MP can intelligently select paths such that all agents avoid each other while sharing the middle volume.

\subsection{Scenario III}

This experiment demonstrates our algorithm's ability to plan for heterogeneous
robot teams. A heterogeneous multi-robot system,
composed of an aerial, a ground-based, and a crane-like robot, must coordinate their motions in
a constrained environment (Fig.~\ref{fig:heter}).

Only CBS-MP and DecoupledPRM were able to solve the heterogenous problem within the 1000 seconds. DecoupledPRM was only successful on 20\% of the trials while
CBS-MP was able to solve in all the trials(Fig.~\ref{fig:het-results}). This demonstrate the effectiveness of the CBS-MP approach and the uniform-time discretization of roadmaps for
finding and resolving conflicts in heterogeneous systems.

\subsection{Scenario IV}
\label{sec::scen4}

Here we evaluate the behavior of CBS-MP and DecoupledPRM as the roadmap scales. Both methods are run with the same roadmaps such that
they query at the same time. CBS-MP is able to find a path with smaller roadmaps than DecoupledPRM, and continues to find better paths as the roadmaps
expands to the point needed by DecoupledPRM to find a solution. DecoupledPRM uses a dynamic-obstacle approach, and it computes the individual paths incrementally. This compromises finding the optimal global solution since computing an optimal agent's path may prevent finding others. Because of that DecoupledPRM will never find a better path than CBS-MP on the same set of roadmaps (Fig.~\ref{fig:fourth-results}).

\section{Conclusions and Future Work}
In this paper, we presented CBS-MP, the extension of CBS to sampling-based motion planning, 
and we apply it to solve the multi-agent motion planning problem. CBS-MP is the 
first decoupled MAMP approach capable of providing a degree of optimality. We validate our 
work in different scenarios to show the strengths of our method to deal with sets of numerous agents, 
sets with high-\dof agents, and sets of heterogeneous agents.

There are several areas to expand this work. The conflict resolution now requires global replanning instead of 
seeking a local repair. The \textit{CT} expands exponentially with the number of conflicts, so useful heuristics
can provide further scalability through efficient construction and exploration of the \textit{CT}.

%\addtolength{\textheight}{-12cm}

% Bibliography ----------------------------------------------------------------%

\bibliographystyle{ieeetr}
\bibliography{robotics.bib}

\end{document}